\documentclass[11pt,a4paper]{article}
\usepackage[hyperref]{acl2018}
\usepackage{times}
\usepackage{latexsym}
\usepackage{amsmath}
\usepackage{amssymb}
\usepackage{graphicx}
\usepackage{scrextend}
\usepackage{url}

\aclfinalcopy % Uncomment this line for the final submission
\title{Detecting Syntactic Change Using a Neural Part-of-Speech Tagger}

\author{William Merrill\thanks{\; Authors (listed in alphabetical order) contributed
equally.}\:\:\thanks{\; Work completed while the author was at Yale University.}\:\:\footnotemark[3]\: \hspace{2em} Gigi Felice Stark\footnotemark[1]\:\:\footnotemark[3]\: \hspace{2em} Robert Frank\footnotemark[3] \\ 
	\footnotemark[3]\: Department of Linguistics, Yale University, New Haven, CT, USA \\
	\footnotemark[2]\: Allen Institute for Artificial Intelligence, Seattle, WA, USA \\
	{\tt first.last@yale.edu}\\}

\date{\today}

\begin{document}
\maketitle
\begin{abstract}

We train a diachronic long short-term memory (LSTM) part-of-speech tagger on a large corpus of American English from the 19th, 20th, and 21st centuries. We analyze the tagger's ability to implicitly learn temporal structure between years, and the extent to which this knowledge can be transferred to date new sentences. The learned year embeddings show a strong linear correlation between their first principal component and time. We show that temporal information encoded in the model can be used to predict novel sentences' years of composition relatively well. Comparisons to a feedforward baseline suggest that the temporal change learned by the LSTM is syntactic rather than purely lexical. Thus, our results suggest that our tagger is implicitly learning to model syntactic change in American English over the course of the 19th, 20th, and early 21st centuries. 

\end{abstract}

\section{Introduction}

We define a diachronic language task as a standard computational linguistic task where the input includes not just text, but also information about when the text was written. In particular, diachronic part-of-speech (POS) tagging is the task of assigning POS tags to a sequence of words dated to a specific year. Our goal is to determine the extent to which such a tagger learns a representation of syntactic change in modern American English.

\par Our method approaches this problem using neural networks, which have seen considerable success in a diverse array of natural language processing tasks over the last few years. Prior work using deep learning methods to analyze language change has focused more on lexical, rather than syntactic, change \citep{Hamilton:2016, dubossarsky:2017, Jo:2017}. One of these works, \citet{Jo:2017},  measured linguistic change by evaluating a language model's perplexity on novel documents from different years.

\par Previous work focusing on syntactic change utilized mathematical simulations rather than empirically trained models. \citet{Niyogi:1995} attempted to build a mathematical model of syntactic change motivated by theories of language contact and acquisition. They found that their model predicted both gradual and sudden changes in a parameterized grammar depending on the properties of the languages in contact. In particular, they used their simulation to study how verb-second (V2) order was gained and lost throughout the history of the French language. For several toy languages, their model found that contact between languages with and without V2 would lead to gradual adoption of V2 syntax by the entire population. 

We use the Corpus of Historical American English (COHA) \cite{COHA}, an LSTM POS tagger, and dimensionality reduction techniques to investigate syntactic change in American English during the 19th through 21st centuries. Our project takes the POS tagging task as a proxy for diachronic syntax modeling and has three main goals:
	\begin{enumerate}
	  \item Assess whether a temporal progression is encoded in the network's learned year embeddings.
	  \item Verify that the represented temporal change reflects syntax rather than simply word frequency.
	  \item Determine whether our model can be used to date novel sentences.
	\end{enumerate}

\section{Materials and Methods}
\subsection{Data}

\subsubsection{Corpus of Historical American English}

The COHA corpus is composed of documents dating from 1810 to 2009 and contains over 400 million words. The genre mix of the texts is balanced in each decade, and includes fiction works, academic papers, newspapers, and popular magazines. Because of computational constraints, we randomly selected 50,000 sentences from each decade for a total of 1,000,000 sentences. We selected an equal number of sentences from each decade to ensure a temporally balanced corpus. We put 90\% of these into a training set and 10\% into a test set. We also cut off all sentences at a maximum length of 50 words. We chose 50 words as our limit to avoid unnecessarily padding a large percentage of sentences. Only 6.95\% of sentences in the full COHA corpus exceeded 50 words.

\par Texts in COHA are annotated with word, lemma, and POS information. The POS labels come in three levels of specificity, with the most specific level containing several thousand POS tags. We used the least specific label for our model, which still had 423 unique POS tags.

\subsubsection{Word Embeddings}

Our model utilized pre-trained 300-dimensional Google News \cite{googlenews} word embeddings that were learned using a standard word2vec architecture. When there was no embedding available for a word in the corpus, we assigned the word an embedding vector drawn from a normal distribution, so that different unknown words would have different embeddings. Due to computational constraints, we only included embeddings for the 600,000 most common words in the vocabulary. Other words were replaced by a special symbol \textit{UNK}.

\subsection{Methods}

\subsubsection{Network Architecture}

\begin{figure*}[h]
   \centering
   \includegraphics[scale=0.45]{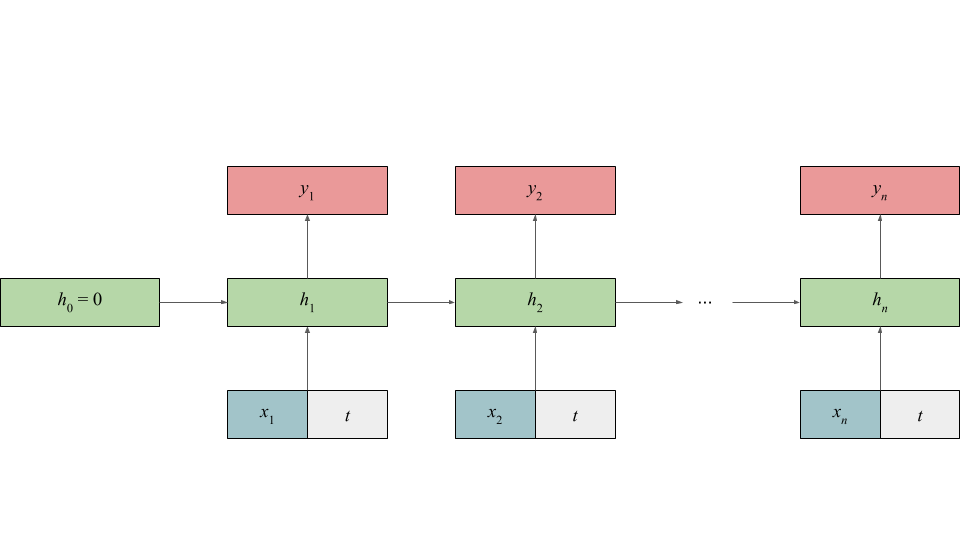}
   \caption{LSTM architecture. The input to the LSTM at each step is the concatenation of the current word's embedding and the corresponding document's year embedding. Each output is the predicted POS tag for the current word.}
   \label{fig:network}
\end{figure*}

We used a single-layer LSTM model.\footnote{\url{https://github.com/viking-sudo-rm/DiachronicPOSTagger}} For a given sentence from a document composed in the year with embedding $t$, the model's input for the $i^{\textrm{th}}$ word in the sentence is the concatenation of the word's embedding $x_i$ and $t$. For example, consider a sentence \textit{hello world!} written in 2000. The input corresponding to \textit{hello} would be the concatenation of the embedding for \textit{hello} and the embedding of the year 2000. A diagram of this architecture can be seen in \autoref{fig:network}.

An interesting feature of our approach is that a single model can learn information about different time frames. Thus, in principle, learning from sentences in any year can inform predictions about sentences in neighboring years. 

\par The word embeddings were loaded statically. In contrast, year embeddings were Xavier-initialized and learned dynamically by our network. Thus, we did not explicitly enforce that the year embeddings should encode any temporal progression.

\par We gave both the word embeddings and year embeddings a dimensionality of 300. We picked the size of our LSTM layer to be 512. Due to the size of our training set and our limited computational resources, we ran our network for just one training epoch. Manual tweaking of the learning rate and batch size revealed that the network's performance was not particularly sensitive to their values. Ultimately, we set the learning rate to 0.001 and the batch size to 100. We did not incorporate dropout or regularization into our model since we did not expect overfitting, as we only trained for a single epoch.

\par In order to calibrate the performance of our LSTM, we trained the following ablation models: 
\begin{itemize}
    \item An LSTM tagger without year input
    \item A feedforward tagger with year input
    \item A feedforward tagger without year input
\end{itemize}

All taggers were trained with identical hyperparameters to the original LSTM. For the feedforward models, the LSTM layer was replaced by a feedforward layer of size 512. The lack of recurrent connections in the feedforward models makes it impossible for these models to consider interactions between words. Thus, these models serve as a baseline that only considers relationships between single words and their POS tags--not syntax.

\subsubsection{Analyzing Year Embeddings}

We aimed to evaluate the extent to which the learned year embeddings encode a temporal trend. We reduced the year embeddings to one-dimensional space using principal component analysis (PCA). We chose PCA because it is a widely used dimensionality reduction technique that requires no hyperparameter tuning. We calculated the correlation between the first principal component of the embeddings and time.

Both the LSTM and feedforward models capture lexical information. However, due to its recurrent connections, the LSTM model is also informed by syntax. To evaluate whether the relationship between years and learned embeddings was due to syntax or simply word choice, we computed the correlation between the first principal component of the embeddings and time for both the LSTM and feedforward models. The difference between the LSTM and feedforward $R^2$ values reflects the degree to which the LSTM's representation of time is informed by syntactic change.

\subsubsection{Temporal Prediction}

We evaluated the ability of our model to predict the years of composition of new sentences. Because this task is difficult for a single sentence, we evaluated model performance at the aggregate level, bucketing test sentences by either year or decade. As the year grouping is much more narrow, model performance when these buckets are used should be worse. We report both year and decade metrics to evaluate the extent to which our model is effective at different levels of specificity. We used our model to compute the perplexity of each sentence in a given bucket at every possible year (1810-2009). We then fit a curve to perplexity as a function of year using locally weighted scatterplot smoothing (LOWESS). These curves provide clear interpretable visualizations that discount extraneous noise. We took the year corresponding to the LOWESS curve's global minimum as the predicted year of composition for the sentences in the bucket.

\par We compared the effectiveness of the LSTM and the feedforward taggers for temporal prediction. For decade buckets, we quantified the predictive power of each model by calculating the average distance across decades between each decade bucket's middle year and predicted year of composition. Similarly, for year buckets, we measured the average distance between the predicted and actual years of composition. For both metrics, the naive baseline model assigns each bucket a predicted year of 1910 (the middle year in the data set), which results in a metric value of 50.0 for both decade and year buckets. 

\section{Results}

\subsection{Tagger Performance}

\begin{table}
\centering
\begin{tabular}{|c|cc|}
	\hline
	 & \textbf{Feedforward} & \textbf{LSTM}\\\hline
	 \textbf{Year} & 82.6 & \textbf{95.5} \\
	 \textbf{No Year} & 77.8 & 95.3 \\\hline
\end{tabular}
\caption{
{Test accuracies for all architectural variants. The networks differed in whether year information was included as input and whether the hidden layer had LSTM or feedforward connections.}}
\label{table:accuracy}
\end{table}

\begin{table}
\centering
\begin{tabular}{|c|cc|}
	\hline
	 & \textbf{Feedforward} & \textbf{LSTM}\\\hline
	 \textbf{Year} & 82.6 & \textbf{95.6} \\
	 \textbf{No Year} & 77.7 & 95.4  \\\hline
\end{tabular}
\caption{
{Training accuracies for all architectural variants.}}
\label{table:trainaccuracy}
\end{table}

Our LSTM POS tagger with year input achieves 95.5\% test accuracy after training for one epoch (see \autoref{table:accuracy}). While we are not focused on achieving state-of-the-art POS tagging performance, this relatively high test accuracy suggests that the tagger is legitimate. The LSTM without year input performed marginally worse with a 95.3\% test accuracy (see \autoref{table:accuracy}). These results suggest that temporal information slightly aids tagging.

\par The feedforward taggers with and without year input had test accuracies of 82.6\% and 77.8\%, respectively (again, see \autoref{table:accuracy}). As feedforward networks, unlike LSTMs, do not take into account relations between words, it makes sense that their POS tagging performance is much lower. Additionally, these results bolster the idea that year input improves tagging performance.

To justify not implementing dropout, regularization, or other techniques to combat overfitting, we  calculated the training set accuracies of each model. For each type of the model, the training set accuracy was comparable to the test set accuracy (see \autoref{table:trainaccuracy}). Thus, our models did not overfit. 

\subsection{Analyzing Year Embeddings}\label{embeddings}

\begin{figure}
   \centering
   \includegraphics[scale=0.5]{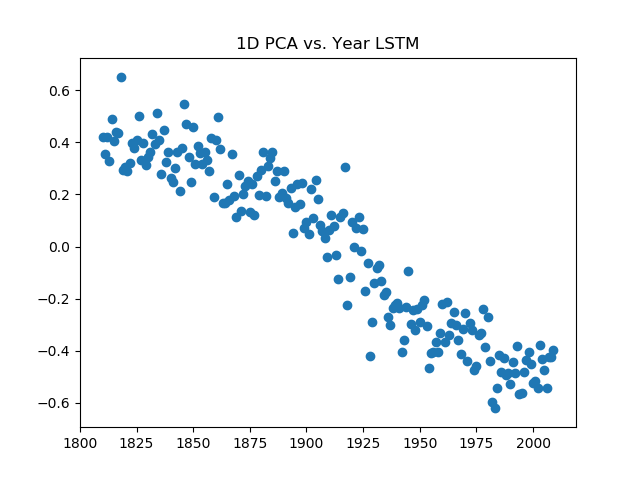}
   \caption{The first principal component of the LSTM year embeddings correlates strongly with time ($R^2=0.89$).}\label{fig:pca_1D}
\end{figure}

\begin{figure}
   \centering
   \includegraphics[scale=0.5]{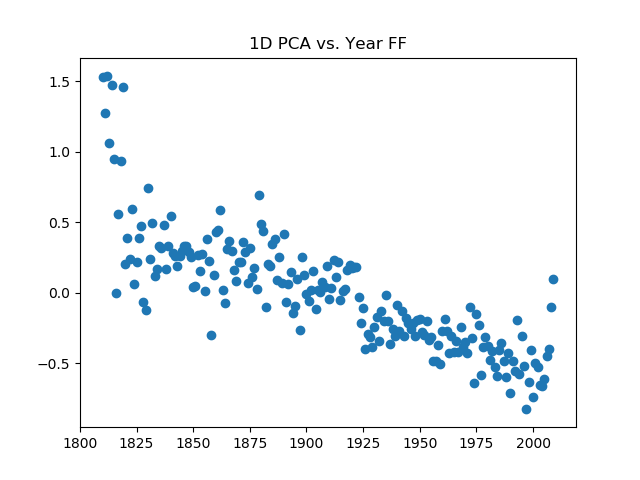}
   \caption{The first principal component of the feedforward year embeddings shows a weaker temporal trend than that of the LSTM ($R^2=0.68$).}\label{fig:pca_FF}
\end{figure}

\par When we plotted the LSTM-learned year embeddings using one-dimensional PCA, a clear linear relationship ($R^2=0.89$) between the years and the first principal component emerged (see \autoref{fig:pca_1D}). These results suggest that the most significant information in the year embeddings encodes the relative position of each year within a chronological sequence. As the first principal component seems to encode temporal information well, we did not see a need to investigate additional principal components. This strong linear correlation suggests that, at the aggregate level, change is monotonic and gradual over time. Even if specific changes do not occur monotonically, the aggregation of these changes allows the network to learn a monotonic representation of change. 

The feedforward network's temporal correlation was weaker ($R^2=0.68$) (see \autoref{fig:pca_FF}). The discrepancy between the LSTM and feedforward $R^2$ values indicates that the LSTM does not only identify effects of lexical change, but also syntactic change. 

\subsection{Temporal Prediction} 

\begin{figure}
   \centering
   \includegraphics[scale=0.4]{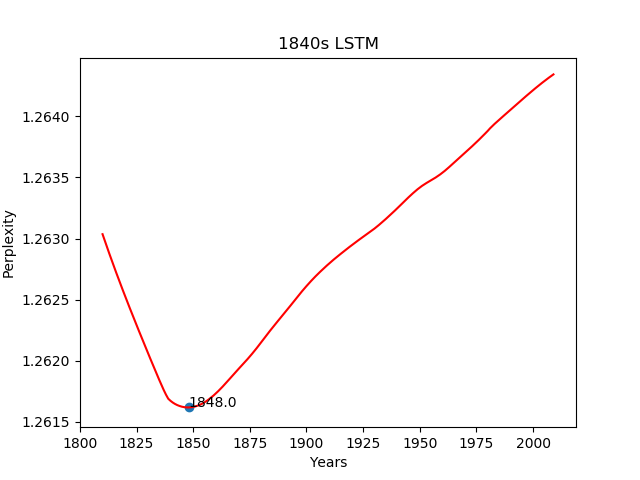}
   \caption{The 1840s LOWESS curve for the LSTM. The year 1848 corresponds to the minimum perplexity.}\label{fig:1850LSTM}
\end{figure}

\begin{figure}
   \centering
   \includegraphics[scale=0.4]{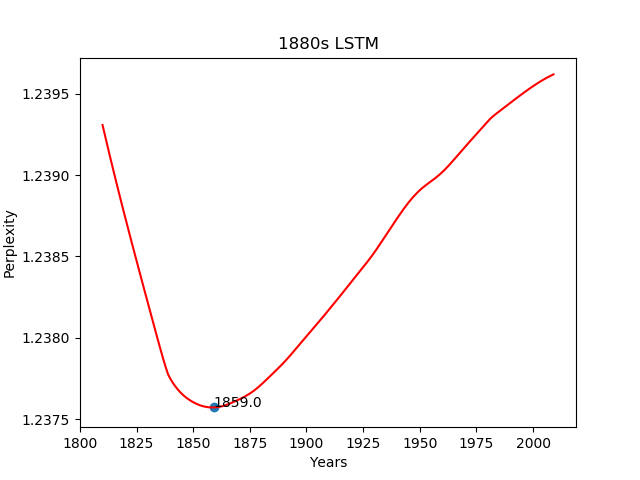}
   \caption{The 1880s LOWESS curve for the LSTM. The LSTM's prediction for this decade is weaker than for the other selected decades. The year corresponding to the minimum perplexity is 1859.}\label{fig:1890LSTM}
\end{figure}

\begin{figure}
   \centering
   \includegraphics[scale=0.4]{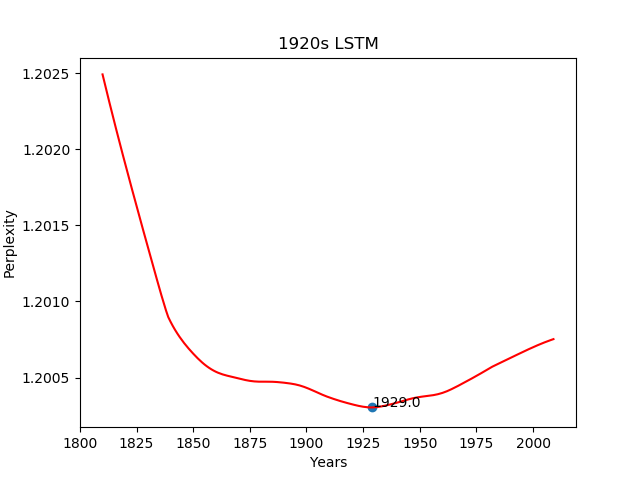}
   \caption{The 1920s LOWESS curve for the LSTM. The year 1929 corresponds to the minimum perplexity.}\label{fig:1940LSTM}
\end{figure}

\begin{figure}
   \centering
   \includegraphics[scale=0.4]{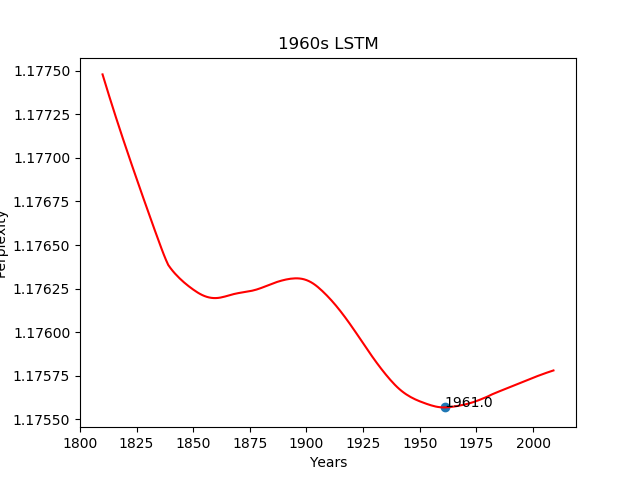}
   \caption{The 1960s LOWESS curve for the LSTM. The year 1961 corresponds to the minimum perplexity.}\label{fig:1950LSTM}
\end{figure}

\begin{figure}
   \centering
   \includegraphics[scale=0.4]{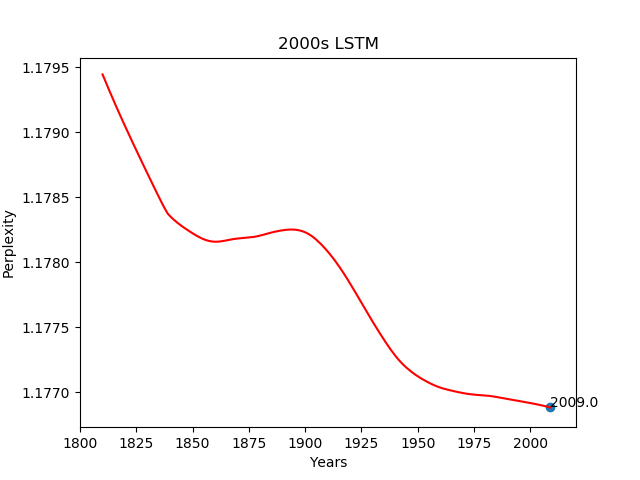}
   \caption{The 2000s LOWESS curve for the LSTM. The year 2009 corresponds to the minimum perplexity.}\label{fig:1970LSTM}
\end{figure}

\begin{figure}
   \centering
   \includegraphics[scale=0.4]{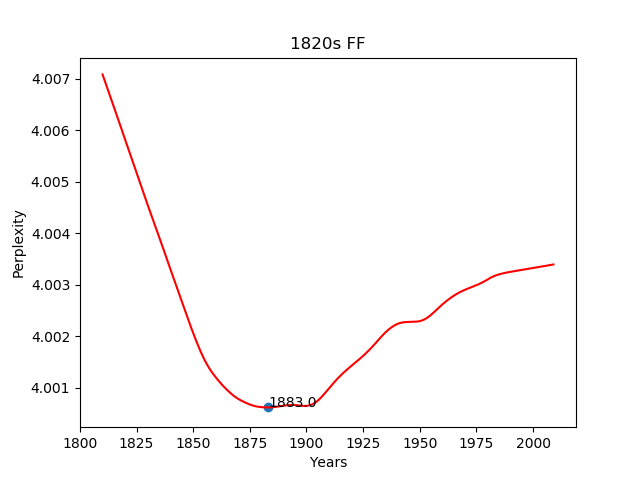}
   \caption{The 1820s LOWESS curve for the feedforward tagger. For this decade, the feedforward network is somewhat off as the year 1883 corresponds to the minimum perplexity. For the feedforward network, there is an evident bias across decades towards the late 1800s and early 1900s, which are the middle years of the data set. }\label{fig:1890FF}
\end{figure}

\begin{figure}
   \centering
   \includegraphics[scale=0.4]{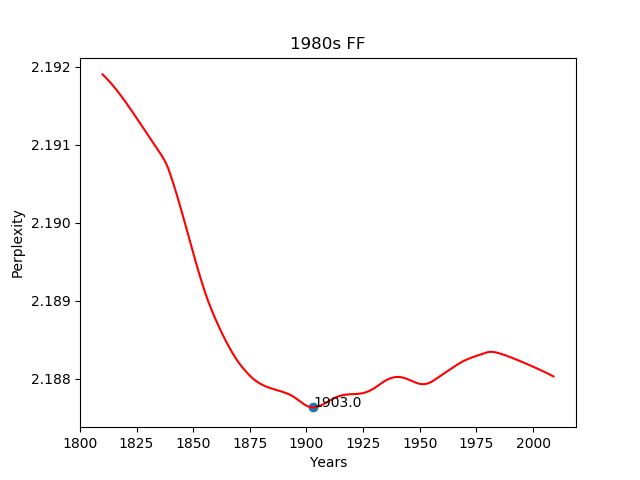}
   \caption{The 1980s LOWESS curve for the feedforward tagger. The feedforward network does not perform well for this decade. The year corresponding to the minimum perplexity, 1903, is somewhat far from the 1980s. Again there is an evident bias in prediction towards the middle years of the data set. }\label{fig:1900FF}
\end{figure}

For both year and decade buckets, the LSTM predicted the years of composition of new sentences much better than the feedforward neural network or the baseline (see \autoref{errorpredictionTable}). We also confirmed our hypothesis that for both types of models the prediction error for decade buckets would be lower than for year buckets. Examples of some of the LSTM perplexity curves can be seen in Figures \ref{fig:1850LSTM}, \ref{fig:1890LSTM}, \ref{fig:1940LSTM}, \ref{fig:1950LSTM}, and \ref{fig:1970LSTM}. Examples of some of the feedforward curves can be seen in Figures \ref{fig:1890FF} and \ref{fig:1900FF}. For all decades, the feedforward year of composition predictions tended to be skewed towards the middle years of the data set (late 1800s and early 1900s). These findings suggest that the representation of syntactic change learned by the LSTM can be leveraged to date new text.

\begin{table}
\centering
\begin{tabular}{|c|ccc|}
	\hline
	 & \textbf{Baseline} & \textbf{Feedforward} & \textbf{LSTM}\\
	 \hline
	  \textbf{Decade} & 50.0 & 26.6 & \textbf{12.5}\\
	  \textbf{Year} & 50.0 & 37.5 & \textbf{21.9}\\
	\hline
\end{tabular}
\caption{
{Average distance between each time period's center and the year that minimizes the perplexity value of the corresponding LOWESS curve. For the decade-level metric, the ``center" is the middle year of the decade (1815 for 1810s). For the year-level metric, the ``center" is the year itself (1803 for 1803).}}
\label{errorpredictionTable}
\end{table}

% \begin{figure*}
% \centering
% \begin{minipage}{.5\textwidth}
%   \centering
%   \includegraphics[width=.4\linewidth]{image1}
%   \captionof{figure}{A figure}
%   \label{fig:test1}
% \end{minipage}%
% \begin{minipage}{.5\textwidth}
%   \centering
%   \includegraphics[width=.4\linewidth]{image1}
%   \captionof{figure}{Another figure}
%   \label{fig:test2}
% \end{minipage}
% \end{figure*}

We examined sample sentences whose years of composition were predicted well. We sampled 1,000 test data set sentences. Of sentences longer than five words, we examined the ten sentences with the smallest errors (distance between the LSTM-predicted year of composition and the actual year of composition). For each of these sentences, we also calculated the error assigned to the sentence by the feedforward model in order to determine the extent to which syntax aided these predictions. These ten sentences are detailed in \autoref{table:example_sentences}. Generally for these sentences, the feedforward error was comparable to or larger than the LSTM error, which suggests that syntactic information improved prediction for these sentences.

Sentence 1 (again, see \autoref{table:example_sentences}) was one of seven sentences whose year of composition was predicted perfectly. This sentence's predicted year of composition was the same as its actual year of composition (1817). It makes sense that this sentence was predicted well since its syntax is qualitatively archaic. For example, it uses the uninflected subjunctive form \textit{shine} whereas modern American English would prefer \textit{shines}. 

\begin{table*}
\centering
\begin{tabular}{|l|cc|cc|}
	\hline
	 & \multicolumn{2}{c}{\textbf{Year}} & \multicolumn{2}{|c|}{\textbf{Error}} \\
	 \textbf{Sentence} & \textbf{Pred} & \textbf{True} & \textbf{FF} & \textbf{LSTM}\\
	\hline
	\parbox{.65\textwidth}{1. it is of great consequence, that we adorn the religion we profess, and that our light shine more and more that we grow in  grace as we advance in years, and that we do not resemble the changing wind or the inconstant wave. \\} & 1817 & 1817 & 86 & 0 \\
	\hline
	\parbox{.65\textwidth}{2. what extenuations or omissions had vitiated his former or recent narrative; how far his actual performances were congenial with the deed which was now to be perpetrated, i knew not. \\ } & 1827 & 1827 & 0 & 0 \\
	\hline
	\parbox{.65\textwidth}{3. that an unlimited power of making gifts could be narrowed down, by any process of reasoning, to the idea of a grant to an indian, a reward loan informer, and much less to a mere sale for money. \\ } & 1833 & 1833 & 16 & 0 \\
	\hline
	\parbox{.65\textwidth}{4. ``amiable, generous, kindhearted woman! thou wert anxious to procure for thy poor, afflicted, aged mother, all the repose which her advanced life seemed to require, to wipe away the tear from her dimmed eye and farrowed cheek, and as far as \\ } & 1817 & 1817 & 10 & 0 \\
	\hline
	\parbox{.65\textwidth}{5. count when shall we meet again? ther. \\ } & 1821 & 1821 & 11 & 0 \\
	\hline
	\parbox{.65\textwidth}{6. in some instances, upon killing them after a full year's deprivation of all nourishment, as much as three gallons of perfectly sweet and fresh water have been found in their bags. \\} & 1838 & 1838 & 21 & 0 \\
	\hline
	\parbox{.65\textwidth}{7. that's a good way to think about me. \\ } & 1996 & 1996 & 12 & 0 \\
	\hline
	\parbox{.65\textwidth}{8. the contention between the wife of abraham and her egyptian handmaid, has already been the subject of animadversion; but although their histories are considerably blended, some features in the character of the latter, and some affecting circumstances of her life, have been hitherto omitted, \\} & 1816 & 1817 & 0 & 1 \\
	\hline
	\parbox{.65\textwidth}{9. but what happens when your erotic adventure is stifled by an unwelcome companion, such as a roommate? masturbating in a UNK situation does pose some problems, but where there is a will, there is a way. \\} & 2003 & 2002 & 22 & 1 \\
	\hline
	\parbox{.65\textwidth}{10. it is UNK in truth we are an united people it is true but we are, family united only for external objects; for our common defence, and for the purpose of a common commerce; sharing, in com mop, the UNK and privations of war \label{hello} \\} & 1826 & 1827 & 3 & 1 \\
	\hline
\end{tabular}
\caption{
{Sentences whose years of composition were predicted best by the LSTM model. The table includes the actual and predicted years of composition, and the feedforward and LSTM error measures.}}
\label{table:example_sentences}
\end{table*}

% https://tex.stackexchange.com/questions/150492/how-to-use-itemize-in-table-environment

\section{Conclusion}

Through our PCA analysis of the year embeddings, we found that the LSTM learned to represent years in a chronological sequence without any biases imposed by initialization or architecture. The LSTM also effectively predicted the year of composition of novel sentences. Relative performance on these tasks indicates that the LSTM learns a stronger representation of time than the feedforward baseline. Therefore, the diachronic knowledge learned by the LSTM must encompass syntactic--not just lexical--change. 

\par One conceptual puzzle with our results is how to reconcile the continuous notion of change represented by our model with the discreteness of natural language grammar. Some theories explain continuous grammatical change by positing that, at any given time, speakers have multiple grammars, or multiple options for syntactic parameters within a grammar \citep{aboh_2015}. The relative probabilities of different options can change gradually, permitting continuous grammatical change. Further work could use similar methods to examine how neural networks represent patterns of change in specific grammatical constructions. This analysis could evaluate the degree to which individual syntactic changes--rather than aggregate measures of change--are continuous.

% \section{Further Work}

% One way to improve our model might be to pick a representation for the year that enforces the temporal ordering between years (we saw that this was learned naturally by our model, but perhaps enforcing it would lead to better results). One way to do this would be representing the year as a continuous value between 0 and 1. Alternatively, one could use a temperature-slider encoding, where a year $y$ is represented by a vector $x$ according to

% \begin{equation}
%     x_i =
%     \begin{cases}
%         1 & \textrm{if} \; i \leq y \\
%         0 & \textrm{otherwise.}
%     \end{cases}
% \end{equation}

% \noindent Enforcing the ordering of years in this way might yield smoother perplexity curves.

% \par Another direction for further study would be experimenting with the architecture. Some notable design choices are the number and size of the LSTM layers, as well as the size of the year embeddings.

% \section *{Acknowledgments}

% We obtained access to COHA through our affiliation with Yale University. Thank you to Kevin Merriman for his help in expediting this process for us.

% TODO: Acknowledge Bob, maybe Claire here as well.

%\bibliography{acl2018}
\bibliography{acl2018}
\bibliographystyle{acl_natbib}

\end{document}